\documentclass[conference]{IEEEtran}
\usepackage{graphicx}
\usepackage{hyperref}
\usepackage{tabularx}
\usepackage{booktabs}
\usepackage{amsmath}

\title{Improving Human-Robot Teaching by Quantifying and Reducing Mental Model Mismatch}

\author{
    \IEEEauthorblockN{Phillip Richter}
    \IEEEauthorblockA{
        \textit{Bielefeld University}\\
        Bielefeld, Germany\\
        p.richter@uni-bielefeld.de
    }
    \and
    \IEEEauthorblockN{Heiko Wersing}
    \IEEEauthorblockA{
        \textit{Honda Research Institute Europe}\\
        Offenbach am Main, Germany\\
        heiko.wersing@honda-ri.de
    }
    \and
    \IEEEauthorblockN{Anna-Lisa Vollmer}
    \IEEEauthorblockA{
        \textit{Bielefeld University}\\
        Bielefeld, Germany\\
        anna-lisa.vollmer@uni-bielefeld.de
    }
}

\begin{document}

\maketitle

\begin{abstract}
The rapid development of artificial intelligence and robotics has had a significant impact on our lives, with intelligent systems increasingly performing tasks traditionally performed by humans. Efficient knowledge transfer requires matching the mental model of the human teacher with the capabilities of the robot learner. This paper introduces the Mental Model Mismatch (MMM) Score, a feedback mechanism designed to quantify and reduce mismatches by aligning human teaching behavior with robot learning behavior. Using Large Language Models (LLMs), we analyze teacher intentions in natural language to generate adaptive feedback. A study with 150 participants teaching a virtual robot to solve a puzzle game shows that intention-based feedback significantly outperforms traditional performance-based feedback or no feedback. The results suggest that intention-based feedback improves instructional outcomes, improves understanding of the robot's learning process and reduces misconceptions. This research addresses a critical gap in human-robot interaction (HRI) by providing a method to quantify and mitigate discrepancies between human mental models and robot capabilities, with the goal of improving robot learning and human teaching effectiveness.
\end{abstract}

\section{Introduction}
The rapid development of artificial intelligence and robotics has significantly changed many aspects of our lives \cite{shaukat2020impact,nitto2017social,cockburn2018impact,webb2019impact}. Intelligent systems are increasingly taking over tasks traditionally performed by humans, from industrial production lines to personal assistants in smartphones \cite{mark2021industrial,hoy2018alexa}. This automation can simplify our lives by handling repetitive or dangerous tasks. However, teaching robots effectively requires robust feedback mechanisms tailored to human-robot teaching scenarios. Feedback plays a crucial role in ensuring that robots learn efficiently while minimizing misunderstandings arising from misaligned mental models \cite{huang2019nonverbal, ramaraj2021robots}.

\begin{figure}[h]
    \centering
    \includegraphics[width=0.4\textwidth]{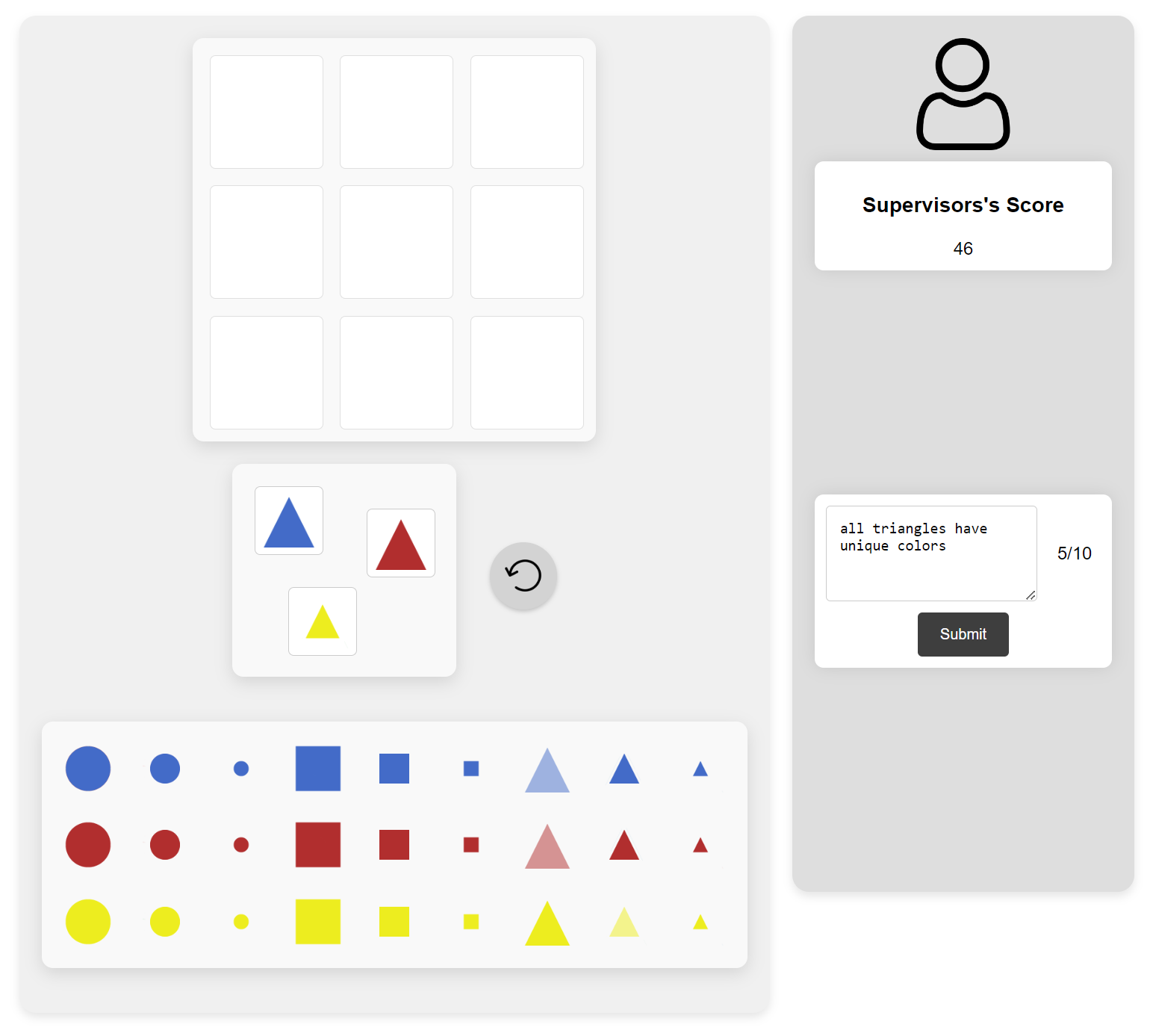}
    \caption{Interactive user interface used to teach the RiddleBot to solve the puzzle game Superdoku. The interface includes a 3x3 grid for the RiddleBot's task, an input interface for tokens, and fields for user input and feedback visualization.}
    \label{fig:superdokugui}
\end{figure}

The interaction between humans and intelligent systems can be complex \cite{soffker2008interaction}. Intelligent systems learn through specific algorithms that differ significantly from human learning behavior \cite{injadat2021machine}. Humans tend to project their own experiences and perceptions onto the system, resulting in a mental model that often does not match the real functioning of the system \cite{ososky2013influence,lee2005human}. This discrepancy, or mental model mismatch, can lead to ineffective teaching methods that fail to take full advantage of the system's learning capabilities, thereby slowing and hindering the teaching process \cite{luebbers2024explainable,hindemith2021robots}.

With this in mind, the research question arises as to whether the inclusion of the human mental model of the intelligent system, explicitly incorporated into the feedback generation process, leads to humans better understanding the architecture of the system and thus being able to teach it more effectively.

The relevance of this research derives from the observation that a high mental model mismatch can lead to frustrating and inefficient interactions between humans and intelligent systems \cite{bansal2019beyond}. Misaligned feedback from the system not only fails to aid teaching but can also be counterproductive \cite{hugo2005semiotics}. An understanding of the human's mental model is critical to creating adaptive feedback mechanisms that respond to user expectations, potentially improving overall interaction quality \cite{priest2002understanding}.

This work presents a novel approach: the \textbf{Mental Model Mismatch (MMM) score}, representing the mismatch between a human's mental model and an intelligent system's architecture. The MMM score provides a systematic way to quantify and reduce this mismatch, aligning human teaching behaviors with robot learning. Using the user's natural language input, we abstract a simplified representation of their mental model to generate adaptive feedback. While this method does not fully capture the complexities of human mental models or robot functionality, it provides a practical approach to simplifying and improving human-robot teaching interactions.

\section{Related Work}

The vision of cooperative intelligence, emphasises the importance of mutual understanding between humans and intelligent systems to achieve synergy towards human-centred values and goals \cite{sendhoff2020cooperative, kruger2017tools}.

Mental models are intrinsic frameworks that human individuals use to interpret, understand, and engage with the world around them. These cognitive constructs are shaped by personal experiences, education, and cultural background, allowing individuals to predict outcomes, make decisions, and solve problems \cite{johnson1983mental}. This concept is influential in cognitive psychology, user experience design, and education \cite{johnson1980mental, villareale2021understanding, boyan2011challenge}.

In HRI, mental models refer to individuals' cognitive representations of intelligent systems, such as robots, including beliefs about their capabilities and operation principles  \cite{tabrez2020survey,nikolaidis2012human}. These models  shape individuals' expectations and interactions with robots, particularly in scenarios where understanding a robot's architecture and functionality is critical for effective task performance \cite{hwang2005role,tabrez2020survey,gervits2018shared}.

Mental models are a key aspect to building understandable robots \cite{hellstrom2018understandable,huang2019enabling}.
The Theory of Mind describes the ability to understand that other individuals have thoughts and feelings that are different from one's own \cite{frith2005theory}. This involves creating mental models of other individuals' mental states in order to predict and explain their behavior \cite{verbrugge2008learning}. Further, the Theory of Mind was identified as a crucial component of the mental model for cooperative interactions and efficient joint work performance \cite{favier2023models, buehler2020theory, frijns2023communication}. Minimizing the mismatch between human and robotic mental models is essential for the comprehension of intelligent agents \cite{hellstrom2018understandable}.

Interactive task learning (ITL) in HRI involves humans teaching new tasks via natural interactions to robots, using techniques such as demonstration, feedback and dialogue \cite{laird2017interactive, amershi2014power}. This approach improves the adaptability and efficiency of robots in dynamic environments by enabling them to learn skills directly from human users, eliminating the need for extensive pre-programming \cite{liu2020skill, booth2022revisiting}.

Recent studies have explored scaffolding techniques, which involve structured guidance during the teaching process, to improve the human teacher’s ability to communicate concepts to robots \cite{liu2020skill}. For example, providing visual cues or simplified explanations of the robot’s learning mechanisms has been shown to reduce frustration and improve task performance \cite{chiyah2020natural}. These techniques highlight the potential for well-designed interfaces and feedback systems to enhance user engagement and teaching efficiency.

Metrics commonly used to evaluate teaching efficiency include task completion rates, concept acquisition rates, and subjective user satisfaction. However, these metrics often fail to address the alignment of user intentions with robot learning outcomes.

Mental models have been identified as a crucial didactic conceptual tool for understanding the process of human teaching in complex subjects like mathematics or physics \cite{greca2002mental}. Successful teaching requires aligning the existing and acquired mental models of the student with the teacher's target mental model. This in turn makes it important for the teacher to observe and estimate the internal mental model of the student to notice any inconsistencies \cite{prediger2017deepening}. A similar approach seems interesting for teaching in HRI.

Recent research has investigated the quantification of teaching efficiency of human teachers in robotic manipulation tasks. It was shown how incorporating the teacher’s understanding significantly improves the efficiency and effectiveness of human teaching \cite{sena2020quantifying}. This finding on teaching efficiency highlights that accurate mental models can help in improving both the teaching process and the robot's learning outcomes.

The complexity of robots makes it challenging for humans to create accurate mental models of their architecture and functions \cite{ramaraj2021robots,rueben2020estimating}. Mismatches between these mental models and a robot's actual capabilities can lead to misunderstandings, reduced efficiency, and even collaborative task failures \cite{chiyah2020natural,ramaraj2021robots,akgun2014hri}.

Feedback mechanisms are central to enhancing the teaching and learning experience in HRI, as they provide users with actionable insights into robot behavior, improving their understanding and engagement. Research shows that feedback mechanisms, such as visual representations or verbal confirmations, help align human teaching intentions with robot learning outcomes, fostering more intuitive and productive interactions \cite{thomaz2008teachable, elor2022human}. For example, interactive teaching setups where robots provide real-time visual feedback have demonstrated significant improvements in user satisfaction and task performance \cite{elor2022human,thomaz2008teachable,huang2019nonverbal,sena2018teaching}.
While performance-based feedback is commonly used, it often neglects the human teacher’s intentions, leading to misaligned teaching outcomes.

Human demonstration, often operationalized through Learning from Demonstration (LfD), is a widely adopted method for teaching robots new tasks \cite{amershi2014power}. These techniques enable robots to learn directly from human-provided examples, reducing the need for extensive pre-programming. However, studies reveal that the success of LfD is significantly influenced by the accuracy of the human teacher’s mental model of the robot’s learning process \cite{sena2020quantifying}. For instance, systems that incorporate real-time user feedback loops—such as showing the robot’s current understanding of a demonstrated concept—yield better outcomes than those relying solely on predefined training protocols \cite{booth2022revisiting}.

The development and refinement of mental models in HRI is significantly influenced by the joint action between the teaching methods of the human and robot feedback mechanisms \cite{le2023bringing,chi2023calibrated}. This dynamic interplay ensures that both human and robot continuously adjust their approaches to achieve better mutual understanding \cite{sano2007multi}. A calibrated approach to teaching, adapting strategies based on robot performance, highlights the role of feedback in refining these strategies \cite{chi2023calibrated}.

A human's mental model of a technical system influences their behavior \cite{carroll1987chapter,jung2016effects} and in teaching/learning interactions, it influences their strategies and intentions in teaching \cite{sheeran2002intention}. 
Human intentions refer to the goals or purposes that individuals want to achieve by their actions \cite{anscombe2000intention}. These intentions can be influenced by various factors such as personal desires, moral values, social norms or extrinsic pressures \cite{bratman1989intention,qian2023role,ajzen1974factors,malhotra2008endogenous}. Understanding human intentions is crucial in fields such as psychology, artificial intelligence and ethics as it helps in predicting behavior, designing human-centric systems and assessing moral responsibility \cite{miller2019explanation,van2022human,manski1990use}. 
Integrating human intentions into robot feedback systems advances the creation of more intuitive and effective HRI \cite{losey2018review}. By employing techniques such as fuzzy variable admittance control and belief space control with intention recognition, the goal is to improve robot responsiveness to human actions and intentions, thereby fostering safer and more efficient cooperation in tasks that require close human-robot collaboration \cite{ying2022human, braun2022belief}. The perception-intention-action cycle helps robots better understand and predict human actions, enhancing trust and safety in collaborative environments \cite{dominguez2023perception}. Additionally, non-verbal feedback from robots, showing predictions of human actions, improves the teaching process and the human's mental model of the robot learner \cite{huang2019nonverbal}.

A clear distinction between intentions and mental models is critical to this discussion. Intentions refer to the specific goals or objectives that guide a human’s teaching actions, such as “teaching the robot to recognize unique shapes” \cite{anscombe2000intention, bratman1989intention}. Mental models, on the other hand, represent the broader cognitive framework through which humans interpret and engage with the robot’s capabilities and learning processes \cite{johnson1983mental, hellstrom2018understandable}.

Addressing discrepancies between individual's mental models and robot's actual functionalities can improve not only efficiency in teaching tasks, but also safety, trust, and overall user experience in HRI \cite{honig2018understanding}. The development of metrics to quantify these discrepancies, could provide a systematic approach to assessing and mitigating these gaps \cite{sena2020quantifying}. Such metrics could pinpoint specific areas where users' understanding diverges from reality,  aiding targeted education and training efforts. Accurate mental models are essential for improving task performance, fostering positive human-robot relationships, and unlocking the full potential of robotic assistance \cite{hellstrom2018understandable}.

In summary, understanding and aligning mental models in HRI is crucial for enhancing collaboration between humans and robots. Addressing the mismatch between human expectations and robot capabilities to incorperate effective feedback mechanisms, should significantly improve interaction efficiency, safety, and user satisfaction. Developing precise metrics to evaluate mental model discrepancies offers a pathway to refine these interactions systematically.

\section{Contribution}
In scenarios where a human teaches a robot to perform a task, a common challenge is that the robot has the physical ability to perform the task, but lacks the concepts and instructions that are necessary to perform it. This knowledge must be provided by the human who knows these concepts and instructions. However, humans often do not fully understand how robots learn or how to teach them effectively, or they may have incorrect assumptions about their learning behavior.

A review of the current state of research on mental models in HRI revealed a gap: There are methods to improve and measure human learning behavior regarding robots, but no clear metric exists for assessing the misalignment between human mental models and robot capabilities.
To address this we present a novel approach: the Mental Model Mismatch Score. This measure aims to quantify the mismatch between an individual's mental model of how a robot learns and the robot's actual learning architecture. To derive the teacher's mental model, we analyze the intention behind their teaching actions, focusing on the intended learning outcomes for the robot as conveyed through natural language by the teacher. This approach leverages Large Language Models (LLMs) to interpret and extract key teaching intentions, providing a mechanism for bridging human expectations with robotic capabilities.
From the known instructions for completing the task, we derive a set of concepts that the robot must learn to successfully complete the task. Feedback from the learning robot can then be used to refine the human's mental model of how to teach the robot. However, this feedback often fails to take into consideration the mismatch in the user's mental model. Incorporating user mental models into the feedback loop through the MMM score not only identifies teaching errors but also provides actionable guidance to improve alignment between user expectations and robot learning outcomes.

In this initial exploratory approach, our goal is to measure the extent of the mental model mismatch and investigate whether incorporating the human teacher's mental model into the feedback mechanism can reduce this mismatch. The human mental model is inherently complex, and here, we simplify it to a basic layer. This simplification allows a simple representation of the mental model to be derived with minimal effort and in real time during the teaching session. Ultimately, our goal is to investigate whether reducing the mental model mismatch can improve the effectiveness of human-robot teaching interactions. 

To ensure our approach is universally applicable across various scenarios, we utilize simple vectors onto which the intentions are mapped. To elaborate, we employ a concept dictionary where the concepts are represented as either true or false. However, other vectors can also be selected. For instance, we could measure how far the taught content deviates from the user’s intended teaching objective. Additionally, the MMM score framework provides a foundation for integrating more advanced semantic representations, enabling its extension to diverse HRI applications.

\section{Methodology}
Our methodology aims to evaluate the alignment between human teaching intentions and what the robot actually learns. To achieve this, we introduce the \textit{Mental Model Mismatch Score} (\(S_d\)). By leveraging Large Language Models (LLMs) for key term extraction and concept alignment, we systematically evaluate the congruence between human intentions and robot capabilities.

\subsection{Concept Representation}

In our framework, \(\mathcal{C}\) denotes the set of all possible concepts that the system can learn. Each concept \(c_i \in \mathcal{C}\) is represented by a binary state: 1 indicates the concept has been learned, while 0 indicates it has not been learned in a given iteration \(\mathcal{I}\). The key terms extracted from the user's instructional content are represented as \(\mathcal{K} = \{k_1, k_2, \ldots, k_o\}\). Each term \(k_j\) corresponds to a potential match with one or more concepts in \(\mathcal{C}\), thus forming the basis for evaluating the alignment between user intentions and the system's learning.

\subsection{Key Term Extraction}

To extract key terms from the user’s instructional content and align them with the system's learning concepts, we utilize an LLM. For this paper we used ChatGPT 4 Turbo \cite{openai2024chatgpt4api}. The process involves three main steps:

\begin{enumerate}
    \item \textbf{Input Preparation}: The intention of the user, provided in natural language along with a dictionary of possible concepts \(\mathcal{C}\) that the robot can learn, are passed to the LLM in a prompt.
    \item \textbf{Key Term Identification}: The LLM processes the intention of the user in real time to identify key terms (\(\mathcal{K}\)) that reflect the user's teaching intentions.
    \item \textbf{Concept Mapping}: The LLM analyzes the identified key terms to determine which concepts in \(\mathcal{C}\) align with the user's intentions. The LLM then outputs a concept dictionary that contains the matched concepts.
\end{enumerate}

\subsection{Semantic Matching Function}

To determine the match between a concept and a key term, we define a matching function \(\mathrm{M}\) that maps pairs of concepts and key terms to binary values:

\[
\mathrm{M}(c_i, \mathcal{K}) = 
\begin{cases} 
1 & \exists k_j \in \mathcal{K} \mid k_j \text{ matches } c_i, \\
0 & \text{otherwise.}
\end{cases}
\]

This function \(\mathrm{M}(c_i, \mathcal{K})\) returns 1 if there is a key term \(k_j\) in \(\mathcal{K}\) that semantically matches the concept \(c_i\), and 0 otherwise. This binary matching provides a clear measure of alignment between user intentions and the system's learning concepts.

\subsection{Mental Model Mismatch Score Calculation}

The Mental Model Mismatch Score (\(S_d\)) quantifies the misalignment between the user’s teaching intentions and the system’s learning outcomes in each iteration. 

\subsubsection{Mental Model Mismatch Score}

For each teaching iteration, the immediate Mental Model Mismatch 
Score (\(S_d\)) is calculated using the formula:

\[
S_d = 1 - \frac{\sum_{i=1}^{|\mathcal{C}|} \mathrm{M}(c_i, \mathcal{K}) \cdot \text{Learned}(c_i)}{N_{\text{learned},d}}
\]

where:
\begin{itemize}
    \item \(\text{Learned}(c_i)\) is a binary indicator that equals 1 if concept \(c_i\) is learned and 0 otherwise.
    \item \(N_{\text{learned},d}\) represents the total number of concepts learned in iteration \(d\).
\end{itemize}

This score reflects the proportion of correctly matched concepts to the total number of concepts learned, thus indicating the alignment of intentions for that iteration.

\subsubsection{Handling Iterations Without Learning}

In cases where no concepts are learned (\(N_{\text{learned},d} = 0\)), the score \(S_d\) would traditionally be undefined. To address this, we set \(S_d\) to 1 for such iterations:

\[
\text{If } N_{\text{learned},d} = 0, \text{then } S_d = 1.
\]

This indicates a complete mismatch or an ineffective teaching iteration, emphasizing the need for improvement in those instances.

\subsubsection{Cumulative Mismatch Score}

To provide a holistic measure of alignment across multiple iterations, we calculate the cumulative mismatch score (\(S_{\text{cum}}\)) which is updated after each iteration using the formula:

\[
S_{\text{cum},d} = \frac{S_{d} + S_{\text{cum},d-1} \cdot (d-1)}{d},
\]

where:
\begin{itemize}
    \item \(S_{\text{cum},d-1}\) is the cumulative mismatch score up to the previous iteration.
    \item \(d\) is the current iteration number.
\end{itemize}

This weighted approach ensures that each iteration equally contributes to the overall assessment, providing a comprehensive measure of the alignment between human intentions and robot learning over time. This is the resulting score which is then used to generate
feedback.

\subsection{Score Dynamics}

To better understand how the score reacts in different scenarios, consider the following cases:

\begin{itemize}
    \item \textbf{No Concepts Learned}: When no concepts are learned (\(N_{\text{learned},d} = 0\)), the score \(S_d\) is set to 1, indicating a complete mismatch or ineffective teaching iteration.
    \item \textbf{Full Alignment}: If all learned concepts perfectly match the user’s intentions (\(N_{\text{learned},d} > 0\) and \(\sum_{i=1}^{|\mathcal{C}|} M(c_i, \mathcal{K}) \cdot \text{Learned}(c_i) = N_{\text{learned},d}\)), the score \(S_d\) is 0, indicating perfect alignment.
    \item \textbf{Partial Alignment}: When some but not all learned concepts match the user’s intentions (\(N_{\text{learned},d} > 0\) and \(0 < \sum_{i=1}^{|\mathcal{C}|} M(c_i, \mathcal{K}) \cdot \text{Learned}(c_i) < N_{\text{learned},d}\)), the score \(S_d\) falls between 0 and 1, reflecting partial alignment.
\end{itemize}

By using these detailed metrics, we can systematically evaluate and improve the interaction between human teaching intentions and robot learning architectures.

\subsection{Practical Application}
In the following, we will provide various calculation examples to illustrate how the MMM score operates. We have taken the examples from our user study, which we will describe later in more detail.
\subsubsection{Example 1: Teaching Unique Colors} \hfill\\
\noindent \textbf{Token Combination:} \\
(Blue, Circle, Small); (Red, Circle, Medium); (Yellow, Triangle, Small)

\noindent \textbf{User Intention:} \\
"I want to show the robot unique colors."

\noindent \textbf{Semantic Matching:} \\
"I want to show the robot unique colors." $\rightarrow$ \textbf{"unique colors"}.

\noindent \textbf{Robot's Learning:} \\
Token Combination has 3 different colors $\rightarrow$ Robot learns \textbf{"unique colors"}.

\noindent \textbf{Mismatch Score:} \\
\textbf{\(S_d\):} 0 (full alignment) $\rightarrow$ \textbf{Positive Feedback}
\subsubsection{Example 2: Partial Alignment in Teaching Unique Attributes} \hfill\\
\noindent \textbf{User Intention:} \\
"different shapes and colors"

\noindent \textbf{Token Combination:} \\
(Blue, Circle, Large); (Red, Circle, Medium); (Yellow, Square, Medium)

\noindent \textbf{Semantic Matching:} \\
"different shapes and colors" $\rightarrow$ \textbf{"unique shapes"} and \textbf{"unique colors"}

\noindent \textbf{Robot's Learning:} \\
Token Combination has 3 different colors but only 2 different shapes $\rightarrow$ Robot only learns \textbf{"unique colors"}

\noindent \textbf{Mismatch Score:} \\
\textbf{\(S_d\):} 0.5 (partial alignment) $\rightarrow$ \textbf{Mixed Feedback}
\subsubsection{Example 3: Misalignment in Teaching Intention and Learning} \hfill\\
\noindent \textbf{User Intention:} \\
"Teach the robot to recognize large tokens."

\noindent \textbf{Token Combination:} \\
(Red, Square, Small); (Blue, Triangle, Small); (Yellow, Circle, Small)

\noindent \textbf{Semantic Matching:} \\
"Teach the robot to recognize large tokens." $\rightarrow$ \textbf{"size: large"}

\noindent \textbf{Robot's Learning:} \\
Token Combination has 3 small tokens $\rightarrow$ Robot learns \textbf{"size: small"}

\noindent \textbf{Mismatch Score:} \\
\textbf{\(S_d\):} 1 (full misalignment) $\rightarrow$ \textbf{Negative Feedback}
\section{User Study}

To evaluate our approach of utilizing the MMM score as feedback from the robot in a human-robot-teaching scenario, we conducted a user study. To initially assess our overarching concept within a teaching context, we opted for a simplified scenario. This decision was made to mitigate potential variables that are not the focus of our evaluation. Consequently, participants do not interact directly with a real-life robot; instead, they engage with a virtual robot through a graphical user interface (GUI). 

\subsection{Teaching Scenario}
In our teaching scenario, a virtual robot is tasked with solving a puzzle game. The virtual robot is capable of solving the game but lacks knowledge of the necessary concepts to accomplish this. Therefore study participants must teach various concepts to the virtual robot.

\subsubsection{Superdoku}
For this scenario we came up with the puzzle game \textit{Superdoku}. A modified version of the puzzle game Soduko. It consists of different tokens that differ in shape, color, and size. Each of these attributes has three options, resulting in a total of \(3^3 = 27\) unique tokens. The game is set up on a 3x3 grid. The goal of the virtual robot is to fill the grid in such a way that each attribute appears only once in any row and column. For example, there should be only one blue token per row and column, just as a digit in Sudoku must be unique within its row or column.

\subsubsection{Demonstration Learning}
To solve the Superdoku puzzle, the virtual robot needs to understand several concepts, such as the rule that only distinct shapes are allowed in each row and column, or the general concept of the color "blue". Participants are expected to teach these concepts to the virtual robot through demonstration. This is achieved through teaching iterations, where the participants places three tokens and presents them to the robot. The robot can learn different concepts from the presented combination of these three tokens.

The virtual robot's learning algorithm is intentionally simple: if the participant places the correct tokens combination that match a concept, the concept becomes enabled for the robot to use. This approach to the robot's learning may be somewhat unrealistic, as the robot is aware that it has just learnt, but our goal is to evaluate the MMM score rather than the virtual robot's learning algorithm.

The combinations of tokens that lead to the learning of specific concepts are predetermined. For example, if a participant places three tokens of different shapes, the concept of "unique shapes" is activated. If three blue tokens are placed, the robot learns the concept of "color: blue".

The human teachers have to understand that the robot is only learning based on the token combination that is shown. No verbal explanation can be directly employed because the virtual robot can only observe. The participants have to find out that shared features among the tokens are used by the robot for learning concepts.

\subsubsection{Teaching iteration}
During each teaching iteration, the participant provides a brief explanation of their intention behind the selected token combination in a small text box. This explanation accompanies the token combination that is presented to the robot for learning. It's important to note that the virtual robot's learning is based solely on the token combination, not on the participant's stated intention. Both the token combination and the intention are then passed to a supervisor to calculate feedback.

The intention is expressed in natural language and is limited to a maximum of 10 words per teaching iteration.

\subsubsection{Supervisor and Feedback Mechanism}
The supervisor serves as a third entity besides the virtual robot and the participant. The supervisor's role is to compute the feedback score based on the token combination placed, the participant's intention and the virtual robot's learning. We chose to introduce an artificial supervisor to make it clear that the feedback calculation is not performed by the virtual robot. This design choice helps prevent participants from assuming that the virtual robot has access to their verbalized intentions.

After each teaching iteration, the supervisor updates the feedback based on the latest information. Additionally, there is an option to display feedback through demonstration by the virtual robot. In this demonstration, the robot fills the 3x3 grid based on its current understanding of the concepts. This feedback method aims to visually represent the virtual robot's learning progress.

\subsection{Participants}
We recruited 150 participants through the online platform Prolific \cite{prolific}. The participants ranged in age from 18 to 70 years, with an average age of 35.23 years. To ensure clear communication, we used a filter to verify that all participants were fluent English speakers. The sex distribution of the participants was 86 females and 63 males. They took part in the study via a self hosted online platform, which was set up and operated on a server specifically for this purpose. 

\subsection{Study Design}
Our study is divided into three test groups, each consisting of 50 participants. All participants are introduced to the study via a tutorial which explains the process of teaching a robot, the mechanics of the Superdoku puzzle game, and the rules and objectives of the puzzle game. After this introduction, the participant is guided through the GUI of the teaching scenario. Afterwards, the actual teaching scenario starts (see Figure \ref{fig:superdokugui}).

The teaching scenario itself is composed of 25 teaching iterations. After every fifth iteration, the virtual robot provides feedback in the form of a demonstration where it fills in the 3x3 grid based on its current knowledge. If a participant successfully teaches the virtual robot all of the required concepts within these 25 iterations, the participant will be notified that the teaching session has been successfully completed. If the participant is unable to teach all of the required concepts within these 25 iterations, the session will also end and the participant will be notified that the session was not fully successful.
Test groups differ in type of feedback provided:

\begin{enumerate}
    \item \textbf{MMM:} Feedback that includes the MMM score from the supervisor, which incorporates the participant's intentions.
    \item \textbf{Performance:} Binary Feedback from the supervisor that does not consider the participant's intentions and only indicates whether teaching occurred in the last iteration.
    \item \textbf{Baseline:} No feedback from the supervisor.
\end{enumerate}
Participants in all test groups are required to state their intentions in every teaching iteration, regardless of whether this information is used for feedback.
Using this study design, we aim to address the following hypotheses:

\begin{itemize}
    \item \textbf{H\textsubscript{1}:} Feedback that takes into account the teacher's intention results in more effective teaching.
    \item \textbf{H\textsubscript{2}:} Feedback that takes into account the teacher's intention leads to a better understanding by the teacher of how to effectively teach the learning system.
\end{itemize}

\section{Results}
The results of our user study indicate a significant improvement in the human-robot teaching performance when feedback incorporates the participant's intentions through the MMM Score. Participants in the MMM group exhibited an average score of 10.52 (\textit{SD} = 2.37), compared to 9.54 (\textit{SD} = 3.42) in the group receiving only performance-based feedback, and 8.88 (\textit{SD} = 3.43) in the control group with no feedback. Table \ref{table:performance_data} shows the average scores and the corresponding percentages of the maximum possible score. The MMM group's score translates to 80.92\% of the maximum possible score, compared to 73.39\% for the performance-based feedback group and 68.31\% for the control test group.

\begin{table}[h]
\caption{Final average scores over all participants (absolute and percentage of maximum score) per test groups. Results demonstrate a statistically significant improvement for the MMM group (\(t(98) = 1.66, p < .05\)).}
\centering
\begin{tabular}{lcc}
\toprule
Category & Absolute Score & Percentage of Maximum (\%) \\ \midrule
MMM & 10.52 & 80.92\% \\
Performance & 9.54 & 73.39\% \\
Control & 8.88 & 68.31\% \\ \bottomrule
\end{tabular}
\label{table:performance_data}
\end{table}

\begin{figure}[h]
  \centering
  \includegraphics[width=\linewidth]{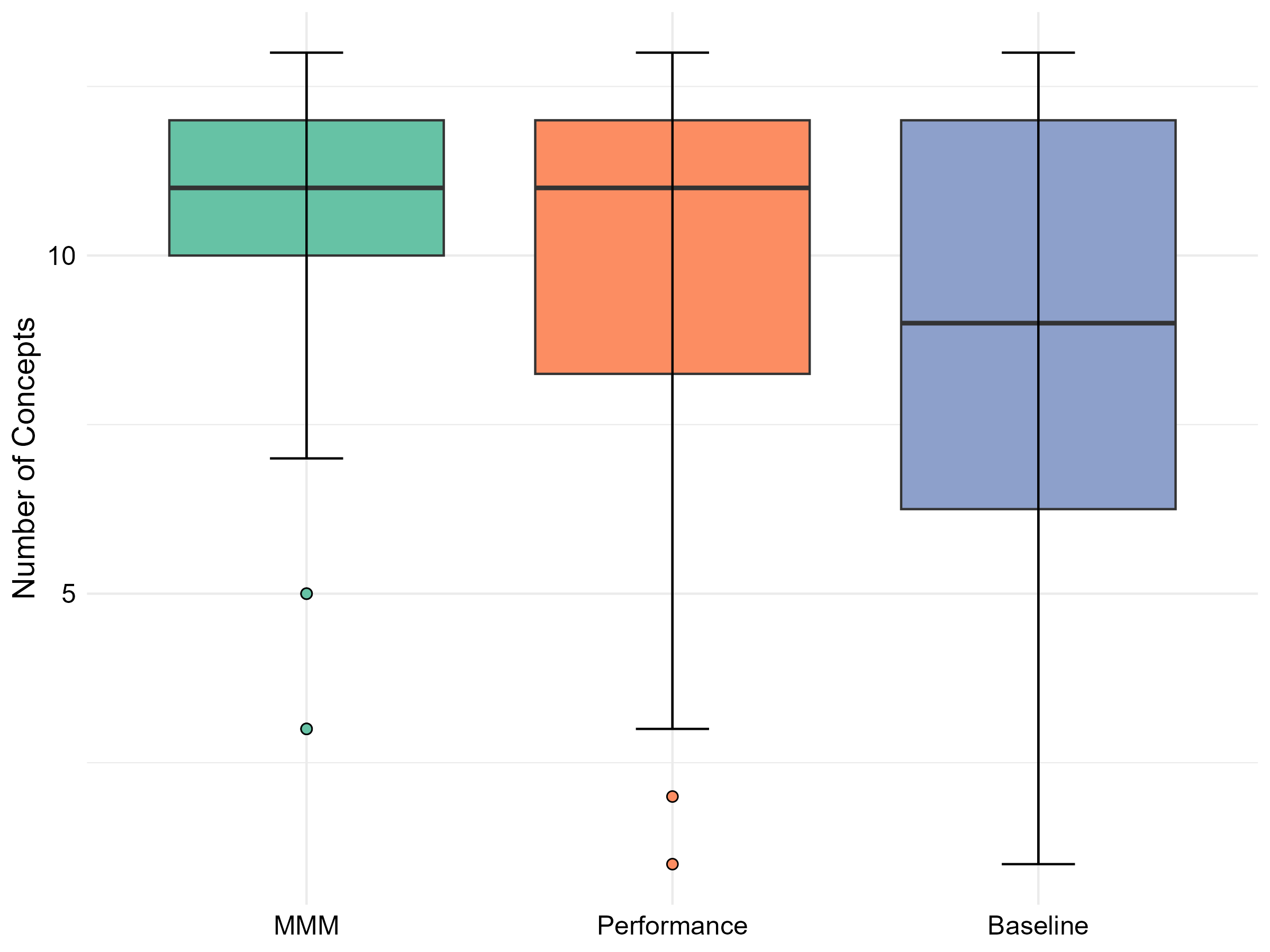}
  \caption{Box plot of scores across the three test groups. The MMM group shows a higher median score and less variability compared to the performance-based feedback and control groups. These results support Hypothesis H1, indicating that feedback incorporating the teacher's intentions significantly improves teaching effectiveness.}
  \label{fig:performance_scores}
\end{figure}

A one-sided independent samples t-test revealed that the MMM group performed significantly better than the performance-based feedback group, \( t(98) = 1.66, p < .05 \), indicating a statistically significant difference at the 95\% confidence level. Additionally, the MMM group performed significantly better than the control group, \( t(98) = 2.78, p = .0032 \). However, there was no significant difference between the performance-based feedback group and the control group, \( t(98) = 0.96, p = .1691 \).

The accompanying box plot (Figure \ref{fig:performance_scores}) provides a visual representation of the distribution of scores across the three test groups. The plot shows that the MMM and performance-based group has a higher median score than the baseline group. The MMM group has less variability compared to the other groups. The scores in the MMM group are tightly clustered around the median, with fewer outliers and a smaller interquartile range (IQR). The performance-based feedback group shows a wider IQR and several outliers, while the control group exhibits the widest range of scores and the largest number of outliers.

\begin{figure}[h]
  \centering
  \includegraphics[width=\linewidth]{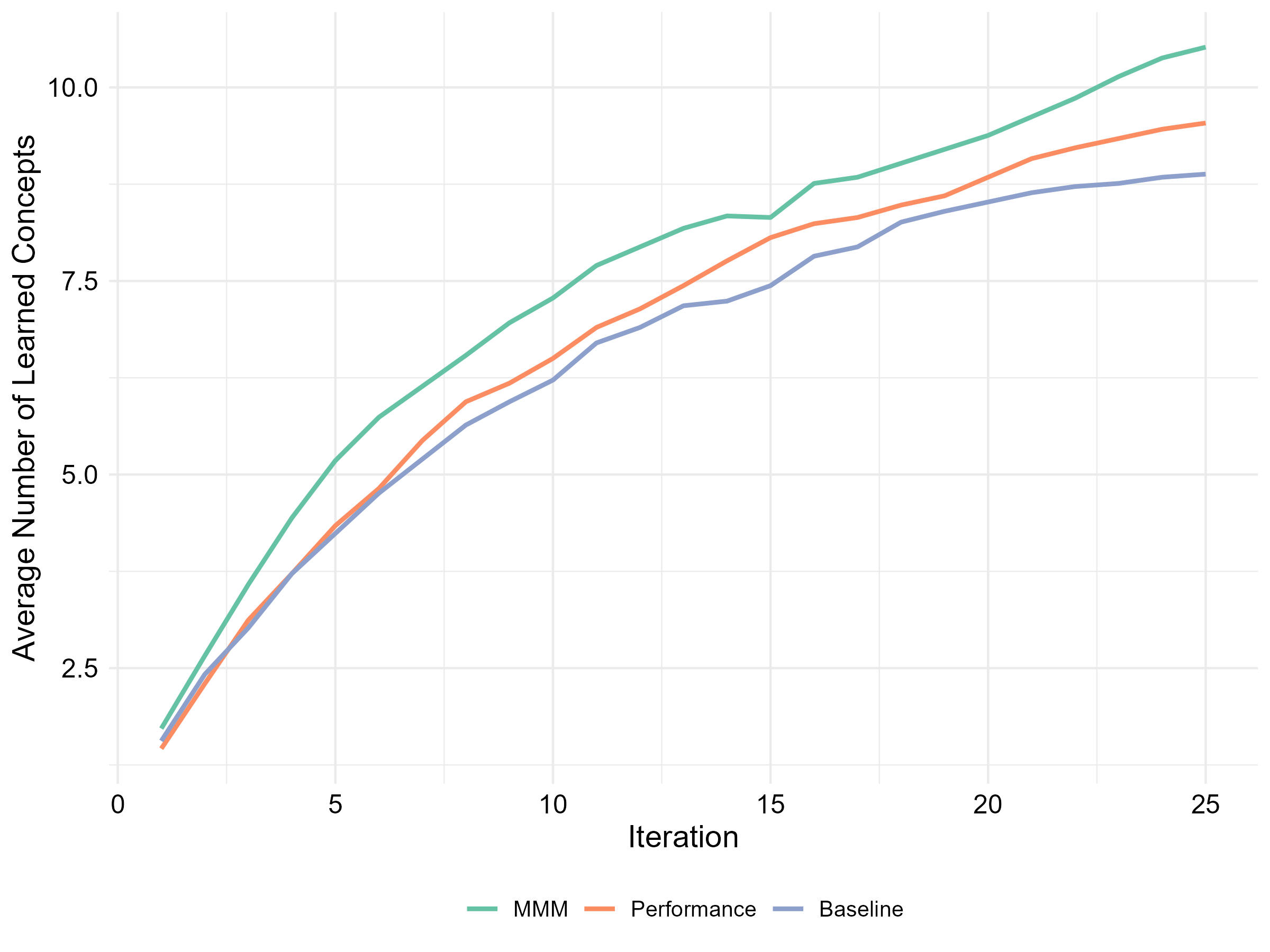}
  \caption{Average number of concepts learned over time across the three test groups. The MMM group demonstrates a steeper learning curve, indicating a higher rate of concept acquisition compared to the performance-based feedback and control groups. For example, participants in the MMM group successfully taught concepts such as "unique colors" and "different shapes" earlier than other groups, demonstrating the benefits of intention-aligned feedback.}
  \label{fig:learned_concepts}
\end{figure}

\begin{table}[h]
\caption{Percentage of participants who teach a concept to the robot per iteration after receiving positive feedback. A consistent trend highlights how intention-aligned feedback sustains participant engagement over multiple iterations, as evidenced by higher percentages in the MMM group.}
\centering
\begin{tabular}{lcc}
\toprule
Iterations & MMM (\%) & Performance (\%) \\ \midrule
1  & 54.37\% & 46.99\% \\
2  & 41.75\% & 38.87\% \\
3  & 37.86\% & 31.10\% \\
4  & 34.95\% & 27.92\% \\
5  & 29.13\% & 26.50\% \\ \bottomrule
\end{tabular}
\label{table:positive_feedback}
\end{table}

\begin{figure}[h]
  \centering
  \includegraphics[width=\linewidth]{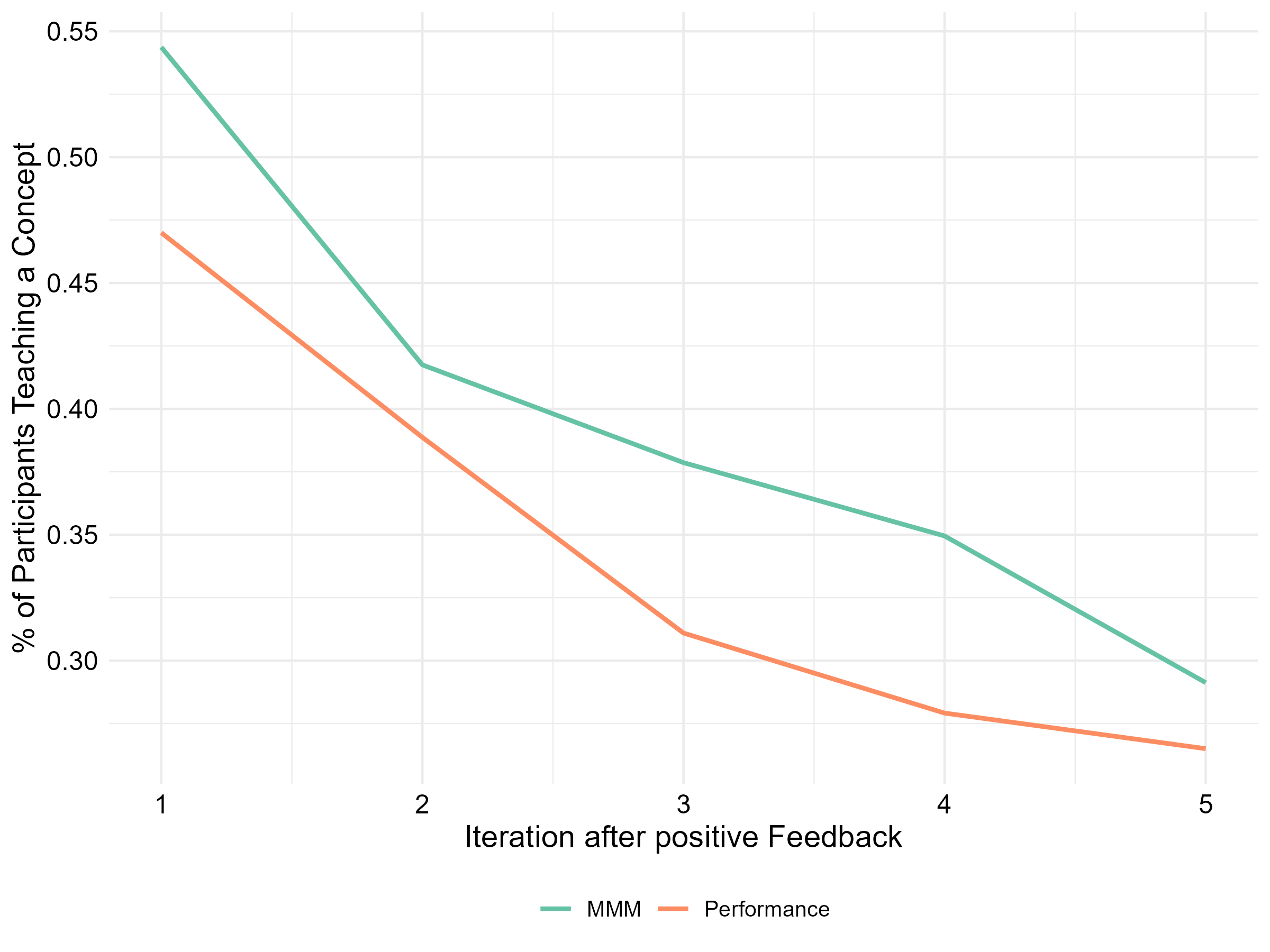}
  \caption{Percentage of participants who thought a concept after receiving positive feedback, across multiple iterations. The MMM group consistently shows higher percentages compared to the performance-based feedback group, further supporting Hypothesis H2.}
  \label{fig:feedback_effects}
\end{figure}

To further understand the impact of feedback on the learning process, we analyzed the number of concepts learned over time by participants in each group. Figure \ref{fig:learned_concepts} presents a line plot that shows the average number of concepts learned over 25 iterations for the MMM group, the performance-based feedback group, and the control group.

The line plot reveals that participants in the MMM group consistently learned more concepts over successive iterations compared to the other groups. In the early iterations, all groups started with a similar number of learned concepts. However, as iterations progressed, the MMM group showed a steeper learning curve.

In addition, we analyzed the impact of positive feedback on participants' subsequent teaching behavior. Table \ref{table:positive_feedback} summarizes the percentage of participants who successfully taught the robot at least one concept across multiple iterations after receiving positive feedback. It is important to note that we focused only on positive feedback in this case, as the score calculation for negative feedback, when participants did not teach a concept in an iteration, is the same for both the MMM and performance-based test groups.

\begin{table}[h]
\caption{Participants' perceptions across the three test groups (1: Strongly disagree, 5: Strongly agree).}
\centering
\begin{tabularx}{\linewidth}{Xccc}
\toprule
Questions & MMM & Perf & Base \\ \midrule
I know from which token combinations the RiddleBot learns anything. & 2.14 & 2.46 & 2.12 \\
I have taught the RiddleBot several concepts. & 3.32 & 3.56 & 2.86 \\
I think the Riddlebot has visual limitations. & 3.7 & 3.6 & 3.82 \\
I know how the supervisor's score is calculated and which of my actions have an influence on it. & 1.78 & 2.24 & - \\
I understand which tokens I have to present the RiddleBot in order for it to learn. & 2.48 & 2.54 & 2.58 \\
I was frustrated while interacting with the RiddleBot. & 4.56 & 4.28 & 4.48 \\
I got helpful feedback from the system. & 1.94 & 2.28 & 2.02 \\
I have a correct mental model about how the RiddleBot learns. & 2.06 & 2.24 & 2.28 \\ \bottomrule
\end{tabularx}
\label{table:likert_data}
\end{table}

Figure \ref{fig:feedback_effects} shows a line plot illustrating these trends, highlighting the percentage of participants who continued to teach more over successive iterations of positive feedback.
The line plot shows the effect of positive feedback over five iterations. For the first iteration, 54.37\% of participants in the MMM group increased their teaching efforts, compared to 46.99\% in the performance-based feedback group. By the second iteration, these percentages dropped to 41.75\% and 38.87\%, respectively. This trend continues with 37.86\% versus 31.10\% in the third iteration, 34.95\% versus 27.92\% in the fourth iteration, and 29.13\% versus 26.50\% in the fifth iteration.

We also collected participant responses using a custom questionnaire with a Likert Scale to gain insights into their perceptions and experiences. The questionnaire focused on various aspects of their understanding and interaction with the robot is summarized in Table \ref{table:likert_data}.
\section{Discussion}
The results of our study support our hypotheses regarding the effectiveness of feedback mechanisms incorporating the teacher's intentions in human-robot teaching.

\subsection{\texorpdfstring{Hypothesis H\textsubscript{1}}{Hypothesis H1}}
Hypothesis H\textsubscript{1} stated that feedback that incorporates the teacher's intentions would lead to more effective teaching. The data support this strongly. Participants in the MMM group, who received intention-based feedback, achieved significantly higher scores than those in the performance-based feedback group and the control group. In particular, the average score for the MMM group was 10.52, or 80.92\% of the maximum possible score, while the performance feedback group scored 9.54 (73.39\%) and the control group scored 8.98 (69.09\%). These results clearly indicate that the inclusion of intention-based feedback leads to more effective teaching outcomes.

The higher median score and lower variability in the MMM group's scores (Figure \ref{fig:performance_scores}) further emphasize the consistency and reliability of intention-based feedback. Participants not only performed better but also achieved more consistent teaching effectiveness, suggesting the feedback mechanism improved their teaching process.

\subsection{\texorpdfstring{Hypothesis H\textsubscript{2}}{Hypothesis H2}}
Hypothesis H\textsubscript{2} suggested that feedback considering the teacher's intentions would improve the teacher's understanding of how to teach the robot effectively. This hypothesis is also supported by our results. The steeper learning curve observed in the MMM group (Figure \ref{fig:learned_concepts}) suggests a higher rate of concept acquisition over successive iterations. Participants in the MMM group consistently learned more concepts, showing that intention-based feedback helped them understand and adapt to the robot's learning more effectively.

These findings directly address the mental model mismatch, as intention-based feedback helps align participants' mental models with the robot's actual learning process. This alignment reduces teaching errors and improves learning outcomes, highlighting the effectiveness of the MMM Score in mitigating cognitive discrepancies during teaching.

Additionally, the effect of positive feedback on subsequent teaching efforts underscores the benefits of intention-based feedback. As shown in Table \ref{table:positive_feedback} and Figure \ref{fig:feedback_effects}, a higher percentage of participants in the MMM group increased their teaching results after receiving positive feedback. This trend persisted across multiple iterations, suggesting that intention-based feedback facilitated immediate teaching success and sustained higher levels of engagement and effort over time.

\subsection{Minimizing Misconceptions}
An important aspect of intention-based feedback in the MMM group is its role in helping to minimize misconceptions about the robot's learning behavior. The MMM Score ensures that participants do not receive positive feedback if they unintentionally teach the robot something that does not match their intended concept. This mechanism helps prevent users from being misled about how the robot learns, as they will not receive positive feedback if they teach a concept that they did not intend to teach. In contrast, participants in the performance-based feedback group receive positive feedback even when they unintentionally teach the robot a different concept than intended. This could reinforce incorrect teaching methods and lead to misunderstandings about the robot's learning behavior. Therefore, the MMM feedback mechanism not only improves teaching effectiveness but also helps to reduce misconceptions about the robot's learning behavior.

\subsection{User Experience}
The results of the custom questionnaire suggest that participants in all groups struggled to understand the learning mechanics of the robot, with low scores for knowing which token combinations the robot learns from (MMM: 2.14, Performance: 2.46, Baseline: 2.12). This shows that understanding how the robot learns was not trivial, as intended. Additionally, high frustration levels were observed in all groups (MMM: 4.56, Performance: 4.28, Baseline: 4.48), indicating that teaching the robot was a challenging task.

The MMM group's lower understanding of the feedback system (1.78 vs. 2.24 in the Performance group) highlights a key limitation of intention-based feedback: its complexity. Future work should focus on enhancing the transparency of these mechanisms, such as by providing real-time visual or verbal explanations of how user intentions are incorporated into the feedback.

Despite these challenges, the better teaching performance of the MMM group may be due to the fact that they are less likely to receive positive feedback when their teaching intentions are not aligned with the learning outcomes, thus preventing them from pursuing incorrect teaching strategies. However, we could not find a better explicit understanding of the system's learning behavior, as illustrated by participants reporting that they do not have a correct mental model of how the robot learns (MMM: 2.06, Performance: 2.24, Baseline: 2.28).

These findings underscore the critical role of feedback methods in human-robot teaching and suggest that sophisticated feedback mechanisms need to be clear to improve teaching effectiveness and user satisfaction.

\subsection{Overall Implications}
The findings from both hypotheses highlight the critical role of feedback mechanisms in human-robot teaching. The significant improvement in performance and concept acquisition in the MMM group reveals that aligning feedback with the teacher's mental model is a promising strategy for improving teaching performance. In contrast, the lower performance of the control group emphasizes the importance of providing feedback to effectively guide the teaching process.

The improvements observed in the MMM group suggest that intention-based feedback mechanisms could play a critical role in scenarios requiring precise human-robot collaboration, such as healthcare, education, and industrial applications. By enabling robots to adapt more intuitively to human instructions, these mechanisms can reduce task completion times, improve user satisfaction, and minimize errors in complex environments.

The variability observed in the performance-based feedback group suggests that feedback based solely on performance indicators may be less intuitive and more difficult for participants to interpret consistently. This variability stands in opposition to the more stable and effective results observed in the MMM group, further supporting the usefulness of intention-based feedback.

While the MMM Score demonstrates significant improvements in short-term teaching tasks, its long-term impact on teaching strategies and user satisfaction remains unclear. Future studies should evaluate whether repeated interactions with intention-based feedback sustain or diminish teaching efficiency and engagement over time. Moreover, testing these mechanisms on physically embodied robots in dynamic environments is critical for validating their applicability beyond controlled experimental setups.

\section{Conclusion}
In this paper, we introduced the MMM score, a novel feedback mechanism designed to align human teaching methods with actual robot learning methods. Our results show that intention-based feedback significantly improves teaching effectiveness. Participants in the MMM group showed higher performance and better understanding of the robot's learning behavior compared to those who received performance-based feedback or no feedback at all.

The MMM Score not only enhances teaching effectiveness but also serves as a foundation for developing more transparent and adaptable human-robot interfaces. By systematically addressing the mental model mismatch, it fosters user trust and provides a scalable framework for teaching diverse robotic systems across various domains.

Our research addresses a critical gap in HRI, where there is a lack of clear metrics for assessing the mismatch between human mental models and robot capabilities. The MMM Score provides a systematic method for quantifying and mitigating these discrepancies, which is necessary for improving efficiency, safety, reliability, and user experience in HRI.

By refining and extending the MMM Score, future research should further improve robot learning performance and human teaching effectiveness. This work contributes to the advancement of intuitive and productive human-robot collaboration and provides practical insights for designing more responsive and intelligent robotic systems.

\section*{Acknowledgments}
Phillip Richter gratefully acknowledges the financial support from Honda Research Institute Europe for the project 'Reducing Mental Model Mismatch for Cooperative Robot Teaching'.

\bibliographystyle{plain}
\bibliography{main}


\end{document}